\documentclass{article} % For LaTeX2e
\usepackage{longcat_style,times}

\usepackage{amsmath,amsfonts,bm}

\def\eqref#1{equation~\ref{#1}}
\def\1{\bm{1}}

\DeclareMathAlphabet{\mathsfit}{\encodingdefault}{\sfdefault}{m}{sl}
\SetMathAlphabet{\mathsfit}{bold}{\encodingdefault}{\sfdefault}{bx}{n}

\usepackage{hyperref}
\usepackage{url}

\usepackage{graphicx}
\usepackage{balance}  % for  \balance command ON LAST PAGE  (only there!)
\usepackage{amsmath}
\usepackage{algpseudocode}
\usepackage{latexsym}
\usepackage{multirow}
\usepackage{multicol}
\usepackage{color}
\usepackage{dashrule}
\usepackage{extarrows}
\usepackage{dsfont}
\usepackage{centernot}
\usepackage{hyperref}
\usepackage{natbib}
\usepackage{fancybox}
\usepackage[utf8]{inputenc} % allow utf-8 input
\usepackage[T1]{fontenc}    % use 8-bit T1 fonts
\usepackage{hyperref}       % hyperlinks
\usepackage{url}            % simple URL typesetting
\usepackage{amsfonts}       % blackboard math symbols
\usepackage{amsmath}
\usepackage{nicefrac}       % compact symbols for 1/2, etc.
\usepackage{microtype}      % microtypography
\usepackage{xcolor} % colors
\usepackage{amsmath}
\usepackage{lipsum}
\usepackage{subcaption}
\usepackage{tabularx}
\usepackage{makecell}
\usepackage{wrapfig}
\usepackage{bm}
\usepackage{multirow}
\usepackage{xcolor} % Load the color package
\usepackage{bbm}
\usepackage{mdframed}
\usepackage{xspace}
\usepackage{longtable,booktabs}
\usepackage{enumitem}
\usepackage{colortbl}
\usepackage[table]{xcolor}
\usepackage{pifont}
\usepackage[most]{tcolorbox}
\usepackage{xcolor}
\usepackage{fancyvrb}
\usepackage{fvextra}
\definecolor{lightgray}{gray}{0.95}
\usepackage{lineno}
\definecolor{darkblue}{rgb}{0, 0, 0.5}

\hypersetup{colorlinks=true, citecolor=darkblue, linkcolor=darkblue, urlcolor=darkblue}

\usepackage{listings}
\usepackage{xcolor}
\usepackage[ruled,vlined]{algorithm2e}
\usepackage{caption}
\usepackage{float} % for H option in algorithm
\usepackage{arydshln}
\usepackage{wrapfig} 
\usepackage{tcolorbox}
\tcbuselibrary{theorems}
\tcbuselibrary{most}
\usepackage[dvipsnames]{xcolor}

\usepackage{hyperref}       % hyperlinks
\usepackage{url}            % simple URL typesetting
\usepackage{booktabs}       % professional-quality tables
\usepackage{amsfonts}       % blackboard math symbols
\usepackage{nicefrac}       % compact symbols for 1/2, etc.
\usepackage{microtype}      % microtypography
\usepackage{xcolor}         % colors
\usepackage{amsmath}
\usepackage{multirow}
\usepackage{multicol}
\usepackage{colortbl}
\usepackage{tcolorbox}
\usepackage{wrapfig}

\usepackage{cleveref}
\usepackage{xspace}

\makeatletter
\DeclareRobustCommand\onedot{\futurelet\@let@token\@onedot}
\def\@onedot{\ifx\@let@token.\else.\null\fi\xspace}

\makeatother

\usepackage{natbib}

\definecolor{light-gray}{gray}{0.6}
\definecolor{front-color}{HTML}{F5FFFA}
\definecolor{Gray}{gray}{0.93}

\definecolor{customTeal}{RGB}{0, 128, 128} 
\definecolor{emphasisColor}{RGB}{255, 0, 0} % Red color for emphasis
\definecolor{oursBlue}{RGB}{51,202,246}

\definecolor{blue1}{HTML}{508AB2}
\definecolor{green2}{HTML}{BFF6BA}

\newtcbtheorem[]{prompt}{Prompt}%
{colback=SeaGreen!10!CornflowerBlue!10,
 colframe=RoyalPurple!55!Aquamarine!100!,
 fonttitle=\bfseries,
 left=.02in, right=.02in, bottom=.02in, top=.02in,
 before upper={\linespread{1.5}\selectfont}}{prompt}

\renewcommand{\shorttitle}{InterleaveThinker: Reinforcing Agentic Interleaved Generation}
\definecolor{darkblue}{rgb}{0, 0, 0.5}
\hypersetup{colorlinks=true, citecolor=darkblue, linkcolor=darkblue, urlcolor=darkblue}

\makeatletter
\renewcommand{\@maketitle}{%
  \vbox{%
    \hsize\textwidth
    \linewidth\hsize
    \vskip -0.5in
    \noindent
    \begin{minipage}{0.99\textwidth}
  \includegraphics[width=0.27\linewidth]{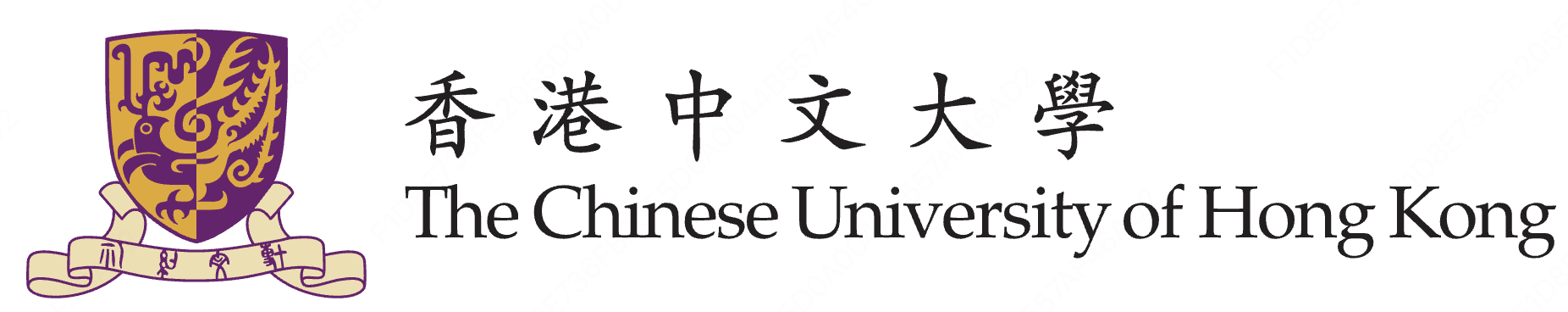}
    \end{minipage}%
    \\
    \rule{\linewidth}{1pt}
    \hspace{0.05\textwidth}%
    \begin{minipage}{0.8\textwidth}
    \end{minipage}

    \centering
    {\LARGE \bfseries\@title\par}
    \vskip 0.1in  % 调整这个值：0.3in=小, 0.5in=中, 0.7in=大
    \def\And{%
      \end{tabular}\hfil\linebreak[0]\hfil%
      \begin{tabular}[t]{c}\bf\rule{\z@}{24\p@}\ignorespaces%
    }
    \def\AND{%
      \end{tabular}\hfil\linebreak[4]\hfil%
      \begin{tabular}[t]{c}\bf\rule{\z@}{24\p@}\ignorespaces%
    }
    \begin{tabular}[t]{c}\bf\rule{\z@}{24\p@}\@author\end{tabular}%
  \vskip 0.05in 
  }
}
\makeatother

\title{InterleaveThinker: \\Reinforcing Agentic Interleaved Generation} 

\makeatletter
\def\@fnsymbol#1{\ensuremath{\ifcase#1\or \dagger\or \ddagger\or
   \mathsection\or \mathparagraph\or \|\or **\or \dagger\dagger
   \or \ddagger\ddagger \else\@ctrerr\fi}}
\makeatother

\newcommand{\homepage}{\raisebox{-1.5pt}{\includegraphics[height=1em]{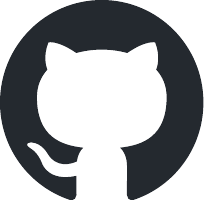}}}
\newcommand{\hfmodel}{\raisebox{-1.5pt}{\includegraphics[height=1em]{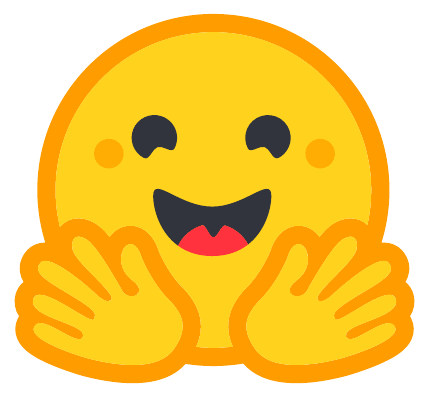}}}

\author{
\begin{tabular}{c}
\textbf{Dian Zheng}$^{1,2}$\textsuperscript{*}
\
\textbf{Harry Lee}
\
\textbf{Manyuan Zhang}$^{2}$\textsuperscript{\dag}
\hspace{0.5em}
\textbf{Kaituo Feng}$^{1}$
\
\textbf{Zoey Guo}$^{3}$
\ 
\textbf{Ray Zhang}$^{1}$
\ 
\textbf{Hongsheng Li}$^{1}$\textsuperscript{\ddag} \\[1ex]
\normalfont $^1$CUHK MMLab \quad $^2$Meituan \quad $^3$CUHK IMIXR\\
{\homepage\ \normalfont 
\texttt{Home: \!\!\!\!\!\url{https://github.com/zhengdian1/InterleaveThinker}}} \\
{\hfmodel\ \normalfont \texttt{HF: \!\!\!\url{https://huggingface.co/InterleaveThinker}}} \\[0.5ex]
{\normalfont\small
\textsuperscript{*}Work done during an internship at Meituan.
\quad
\textsuperscript{\dag}Project Leader.
\quad
\textsuperscript{\ddag}Corresponding Author.}
\end{tabular}
}

\begin{document}
\maketitle

\begin{center}
    \vspace{-2mm}
    \includegraphics[width=0.98\linewidth]{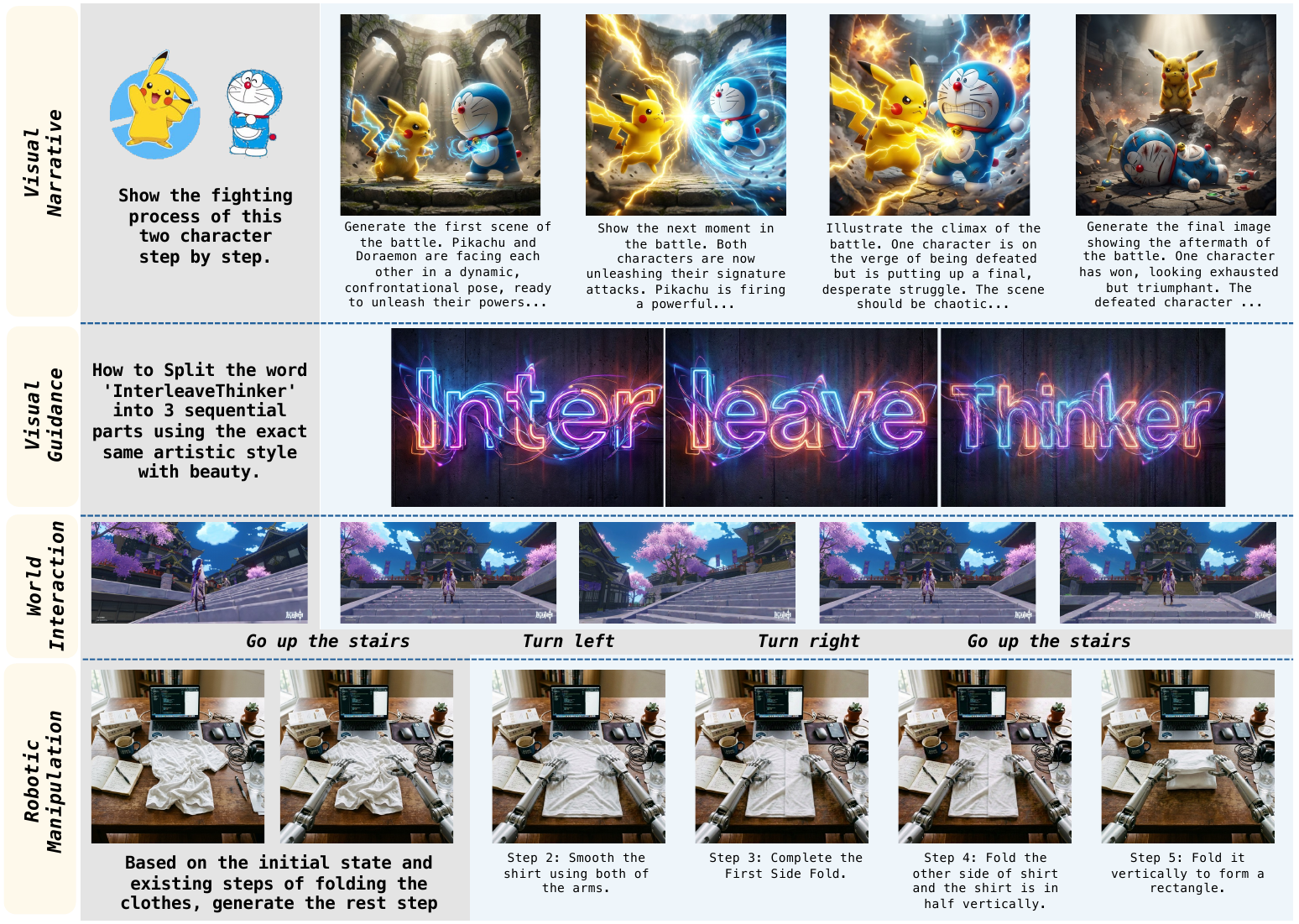}
    \vspace{-2mm}
    \captionof{figure}{\small{Capabilities of InterleaveThinker, consisting of interleaved generation with various types inputs, real-world action interaction, and robotic manipulation.} \textcolor{gray}{Gray}: inputs, \textcolor{blue}{blue}: outputs.}
    \label{fig:teaser}
    % \vspace{-2mm}
\end{center}

\begin{abstract}
Recent image generators have demonstrated impressive photorealism and instruction-following capabilities in single-image generation and editing. However, constrained by their architectures, they cannot achieve interleaved generation (text-image sequence), which has crucial applications in visual narratives, guidance, and embodied manipulation. Even the latest open-source Unified Multimodal Models (UMMs) exhibit limited performance in this regard. In this paper, we introduce InterleaveThinker, the first multi-agent pipeline designed to endow any existing image generator with interleaved generation capabilities. Specifically, we employ a planner agent to organize the image-text input sequence, instructing the image generator on the required execution at each step. Subsequently, we introduce a critic agent to evaluate the generator's outputs, identify samples that deviate from the planned instructions, and refine the instructions for regeneration. To implement this pipeline, we construct the Interleave-Planner-SFT-80k and Interleave-Critic-SFT-112k to perform a format cold-start. Then we develop Interleave-Critic-RL-13k to reinforce the step-wise instruction correction capability within a generation trajectory using GRPO. Since a single interleaved generation trajectory may involve over 25 generator calls, optimizing the entire trajectory is computationally impractical. Therefore, we propose accuracy reward and step-wise reward, allowing single-step RL to effectively guide the entire generation trajectory. The results show that InterleaveThinker improves performance across various image generators. On interleaved generation benchmarks, it achieves performance comparable to Nano Banana and GPT-5. Surprisingly, it also significantly enhances the base model on reasoning-based benchmarks; for example, on 4-step FLUX.2-klein, we observe substantial gains on WISE (from 0.47 to 0.73) and RISE (from 13.3 to 28.9)
\end{abstract}

\section{Introduction}
Recent advancements in image generation and editing have demonstrated remarkable photorealism and instruction-following capabilities. However, these models~\cite{flux2024,flux2,qwenimage,longcatimage,sd3,wang2025ovis} are fundamentally designed for single-image generation/editing. In real-world applications, there is a growing demand for interleaved generation, a workflow that takes an interleaved text and image sequence as input and outputs a coherent, multi-step sequence of text and images. This capability holds crucial value for visual narratives, guidance, and embodied manipulation. Unfortunately, constrained by their inherent image-only output architectures, existing image generators cannot natively achieve this, leaving a significant gap between single-image synthesis and complex sequential generation.

\begin{figure}
    \centering
    \includegraphics[width=0.98\linewidth]{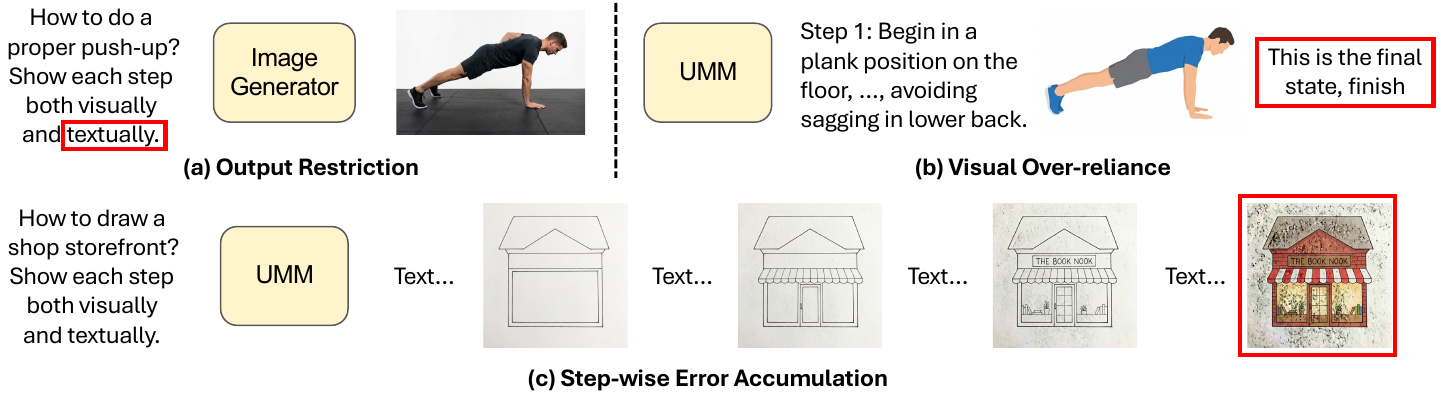}
    \vspace{-2mm}
    \caption{\small\textbf{Problems in image generator and UMM for interleaved generation}. Highlight in \textcolor{red}{red} boxes.}
    \label{fig:problem}
    \vspace{-2mm}
\end{figure}

The emergence of Unified Multimodal Models (UMMs)~\cite{januspro,cui2025emu3_5,nanobanana,bagel,wu2025omnigen2,cao2025hunyuanimage,aia} offers a potential solution, as their architectures naturally support interleaved text and image generation. However, because they generate sequences step-by-step based on preceding images, UMMs suffer from two critical problems in long-horizon tasks: 1) \textbf{Visual over-reliance}. As shown in Fig~\ref{fig:problem}(b), when generating a repetitive action sequence like a push-up, the model might stop at an intermediate state that visually resembles the final goal. 2) \textbf{Step-wise error accumulation}. As current UMMs have not yet achieved a stable ``aha-moment'' for self-correction, a slight degradation in early image quality compounds step-by-step, eventually ruining the final output, as shown in Fig~\ref{fig:problem}(c).

In this paper, we propose InterleaveThinker, the first multi-agent framework that endows any fixed image generator with strong interleaved generation capabilities. The core motivation for this multi-agent design is to eradicate visual over-reliance and resolve step-wise error accumulation through an explicit correction mechanism. If a single VLM alternates between planning and evaluating generated images, it becomes overly conditioned on intermediate visual states. This causes the model to lose sight of the global objective and myopically react to local visual feedback, inevitably leading to step-wise error accumulation. To fundamentally resolve this, InterleaveThinker employs a Planner agent to predict the entire sequence of instructions upfront. This completely bypasses visual over-reliance by blocking intermediate feedback. To monitor the subsequent execution, a Critic agent then evaluates the step outputs, identifies deviations from the initial instructions, and refines prompts for regeneration, ensuring strict adherence to the overall trajectory without updating the generator.

A primary challenge in implementing this multi-agent pipeline is the absence of tailored training data. To address this, we first curate a comprehensive prompt list spanning diverse interleaved generation tasks and scenarios, including embodied manipulation, art, storytelling, image description, workflows, daily life, science, and professional skills. Using these prompts, we iteratively employ advanced models (Gemini 2.5 Pro and Nano Banana Pro) to generate detailed agentic trajectories. To guarantee high-quality supervision, we implement a rigorous data filtering pipeline (detailed in Sec~\ref{subsec:data}). Ultimately, this process yields three high-quality datasets: Interleave-Planner-SFT-80k and Interleave-Critic-SFT-112k to enable the multi-agent format cold-start, alongside Interleave-Critic-RL-13k to reinforce the critic's step-wise correction capabilities using GRPO. Note that one interleaved trajectory can involve over 25 generator calls, so optimizing the entire trajectory end-to-end is computationally impractical. To resolve this, we design a dual-reward strategy comprising an accuracy reward and a step-wise reward. This formulation achieves trajectory-level alignment through efficient single-step RL, drastically reducing computational costs.

To validate the universal applicability of InterleaveThinker, we evaluate the pipeline across multiple off-the-shelf image generators, observing consistent performance gains. As a representative default, we adopt the 4-step FLUX.2-klein to minimize long-horizon latency. Under this setup, our approach significantly surpasses existing open-source UMMs on rigorous interleaved generation benchmarks, achieving performance comparable to the proprietary Nano Banana and GPT-5. Surprisingly, beyond interleaved generation, our framework also significantly enhances the base model on reasoning-based benchmarks. Specifically, we observe substantial improvements on the WISE benchmark (increasing from 0.47 to 0.73) and the RISE benchmark (leaping from 13.3 to 28.9). These results highlight the immense potential of multi-agent collaboration in unlocking complex, sequential reasoning and generation capabilities for existing image models.

In summary, our main contributions are as follows:
\begin{itemize}[leftmargin=1.2em, itemsep=0pt, topsep=0pt, parsep=0pt]
    \item We propose InterleaveThinker, the first multi-agent framework to endow any fixed image generator with strong interleaved generation capabilities. By introducing a Planner-Gen-Critic workflow, it effectively resolves visual over-reliance and step-wise error accumulation in UMM.
    \item To support training, we build a dedicated data pipeline to construct interleaved generation data across diverse scenarios, resulting in three high-quality datasets: Interleave-Planner-SFT-80k, Interleave-Critic-SFT-112k, and Interleave-Critic-RL-13k. In addition, we design a novel dual-reward strategy that achieves trajectory-level alignment through efficient single-step RL via GRPO, drastically reducing computational costs.
    \item Extensive experiments validate the effectiveness and universal applicability of our proposed InterleaveThinker. For example, using 4-step FLUX.2-klein as generator, we not only surpasses existing open-source UMMs on interleaved generation, but also significantly improves the base model on reasoning benchmarks, increasing WISE from 0.47 to 0.73 and RISE from 13.3 to 28.9.
\end{itemize}

\section{Related Works}
\label{sec:related}

\subsection{Unified Image Generation and Editing Model}
Recent advancements in diffusion~\cite{ddpm,vae,controlnet} and autoregressive models~\cite{Qwen2.5-VL,yang2025qwen3} have significantly elevated the photorealism and instruction-following capabilities of image generation models~\cite{flux2024,sd3,podell2023sdxl,qwenimage,gen-searcher}. Building upon these foundational architectures, researchers have developed robust image editing models~\cite{step1x,qwenimage,longcatimage,glm_image,editthinker,zimage,flux2,nanopro}. Crucially, these models preserve their strong text-to-image generation capabilities. Given this dual functionality, we refer to them as unified image generation and editing models (``image generators'' for short), which serve as the base model for our framework. However, their inherent architectures restrict them to interleaved generation, and our work seeks to bridge this gap by retrofitting frozen image generators with robust interleaved generation capabilities.

\subsection{Unified Multimodal Models and Interleaved Generation}
Recently Unified Multimodal Models (UMMs)~\cite{bagel, januspro, longcatnext, wu2025omnigen2, wang2024emu3, cui2025emu3_5, cao2025hunyuanimage,uniedit,aia} have emerged as a promising paradigm. UMMs natively support interleaved generation by modeling text and visual tokens within a unified framework. Despite their architectural advantages, UMMs struggle with long-horizon tasks due to two fundamental issues. First, they suffer from \textit{visual over-reliance}: because they condition heavily on immediately preceding visual states, they frequently halt at intermediate states that superficially resemble the final goal. Second, without a robust self-correction mechanism, minor degradations in early steps lead to severe \textit{step-wise error accumulation}, eventually ruining the final output. DuoGen~\cite{duogen} simulates UMM by jointly tuning a VLM and a video generator. Despite improved performance, it suffers from visual over-reliance and is incompatible with arbitrary image generators. InterleaveThinker overcomes these limitations by decoupling planning and generation, preventing myopic reactions to local visual feedback.

\subsection{Agentic Reinforcement Learning}
Agentic reinforcement learning (RL) has recently emerged as an effective paradigm for training LLMs and VLMs to perform multi-agent, multi-step reasoning and long-horizon tool interaction~\cite{arpo, vds, insight_plus,self-refine,shinn2023reflexion}. In the visual generation domain, researchers have begun adapting agentic RL to enhance output quality and controllability. Gen-Searcher~\cite{gen-searcher} trains search agents to guide knowledge-intensive image generation, \cite{wang2024genartist,yang2024idea2img,reflectdit,reflectionflow,yin2025reasonedit} explore multi-turn refinement for image generation/editing and \cite{editthinker,li2026thinkrl} further employs RL to it. Despite these promising explorations, applying multi-agent RL to long-horizon interleaved generation remains unexplored.

\begin{figure}
    \centering
    \includegraphics[width=0.98\linewidth]{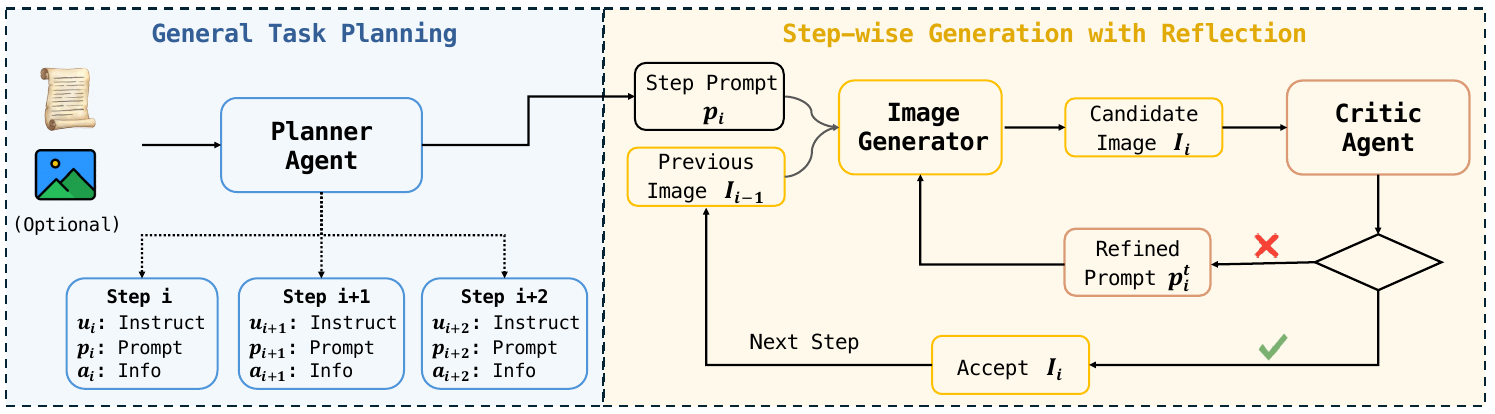}
    \vspace{-2mm}
    \caption{\small\textbf{Overview of InterleaveThinker}. t means the refinement iterations. Fig~\ref{fig:work_exp} for inference example.}
    \label{fig:pipeline}
    \vspace{-2mm}
\end{figure}

\begin{figure}[h]
    \centering
    \includegraphics[width=1.0\linewidth]{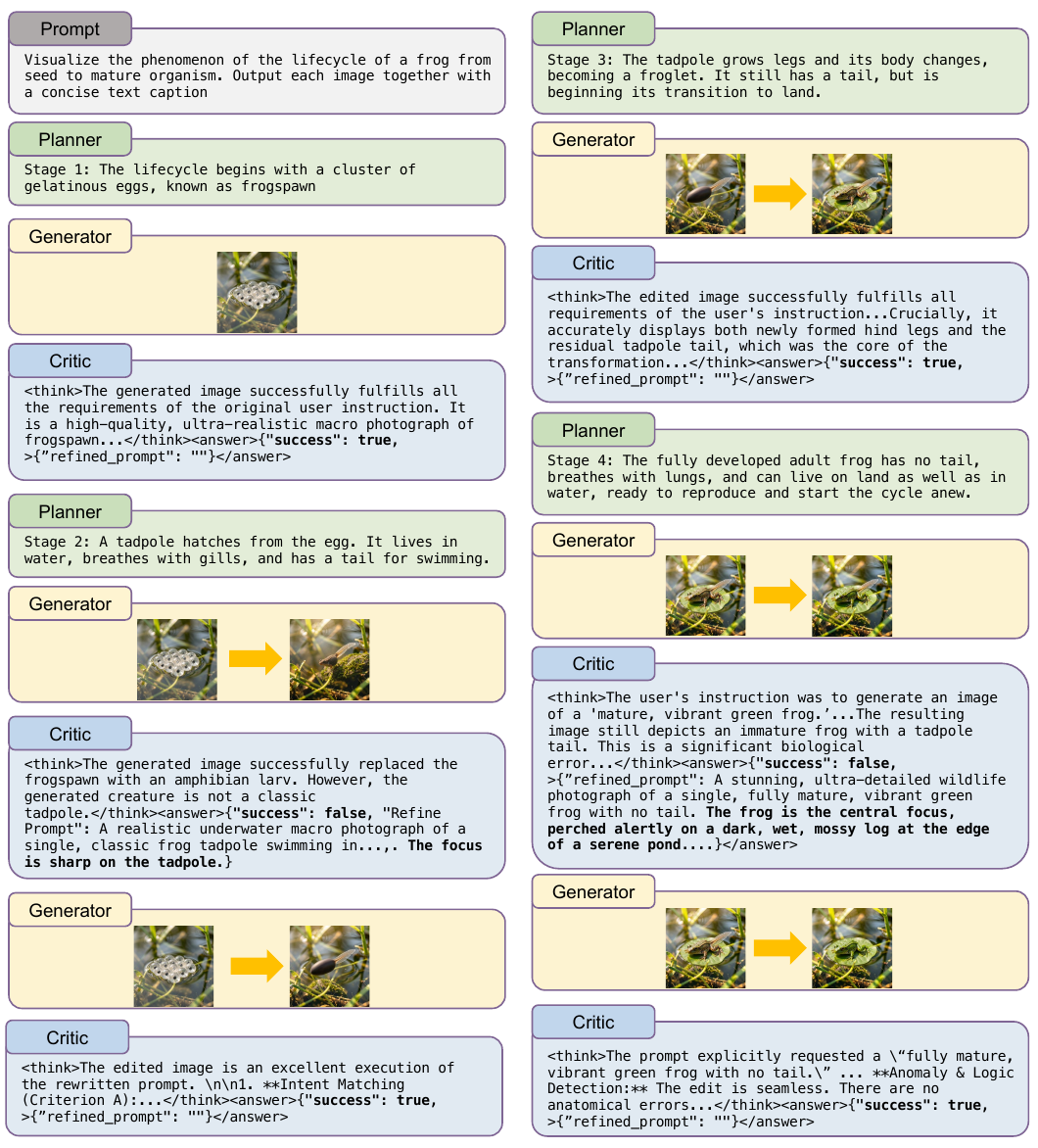}
    \vspace{-6mm}
    \caption{\small\textbf{The working flow of InterleaveThinker}.}
    \label{fig:work_exp}
    \vspace{-6mm}
\end{figure}

\section{InterleaveThinker}
\label{sec:method}

To endow existing frozen image generators with robust interleaved generation capabilities, and address the visual over-reliance, step-wise error accumulation problem in UMM. We propose InterleaveThinker, a universal multi-agent framework. We show our multi-agent workflow, data construction pipeline, and training scheme below.

\subsection{Multi-Agent Pipeline}
\label{subsec:workflow}
As shown in Fig~\ref{fig:pipeline}, we formulate a progressive, closed-loop pipeline comprising three core modules: a \textit{Planner}, a \textit{Critic}, and a \textit{Generator} (Any generator that \textbf{both handles image generation and editing}, such as FLUX.2-klein~\cite{flux2}, Qwen-image-Edit~\cite{qwenimage}). The framework decomposes the complex interleaved generation process into a step-wise execution plan, incorporating self-correction mechanism to ensure high-fidelity generation and editing. Let $S$ denotes the input interleaved sequence of images and text. The overall pipeline operates through the following formalized stages:

\textbf{1. Planner:} 
The Planner is responsible for analyzing the input sequence $S$ and translating it into an $N$-step execution plan. For each step $i \in \{1, \dots, N\}$, the Planner generates a step instruction $u_i$, a model-friendly initial prompt $p_i$ adapted from $u_i$, and an auxiliary text $a_i$, which provides supplementary knowledge-based elaboration required for specific image generation tasks. The planning process is formulated as:
\begin{equation}
   \left\{ \left(u_i, p_i, a_i \right) \right\}_{i=1}^N = \texttt{Planner}(S).
\end{equation}

\textbf{2. Generator:} 
At step $i$ and refinement iteration $t \in \{1, T_{max}\}$, the Generator takes the current refined prompt $r_i^t$ ($r_i^0$=$p_i$) and the image from the previous step $I_{i-1}$ to produce the current image $I_i^t$:
\begin{equation}
    I_i^t = \texttt{Generator}\left(r_i^t, I_{i-1}\right).
\end{equation}
\textit{Note: For the initial generation step i=1, where no prior visual context exists, $I_0$ is defined as $\emptyset$.}

\textbf{3. Critic:} 
To ensure the generated output $I_i$ strictly aligns with the intended instruction $p_i$, we introduce a Critic module that provides quantitative feedback and prompt optimization. At iteration $t$ of step $i$, the Critic evaluates the transition from the pre-execution image $I_{i-1}$ to the post-execution image $I_i^t$. It takes the initial prompt $p_i$ and the current refined prompt $r_i^t$ as textual conditions. The Critic outputs a binary judgment $j_i^t$, a newly refined prompt $r_i^{t+1}$ for the next iteration, and a reasoning process $R_i^t$:
\begin{equation}
    \left(j_i^t, r_i^{t+1}, R_i^t \right) = \texttt{Critic}\left(I_{i-1}, I_i^t, p_i, r_i^t\right),
\end{equation}
\textit{Note: For the initial step i=1, $I_0$ is set as a blank white image to maintain input consistency.}

This generation-evaluation loop (Stage 2$\leftrightarrow$3) iterates until a positive execution judgment (\texttt{True}) is obtained, or a maximum number of iterations $T_{max}$ is reached. Upon satisfaction, the pipeline finalizes $I_i$ and $a_i$, appends them to the output sequence, and proceeds to step $i+1$. We also show a comprehensieve workflow examples in Fig~\ref{fig:work_exp}.

\begin{figure}
    \centering
    \includegraphics[width=1.0\linewidth]{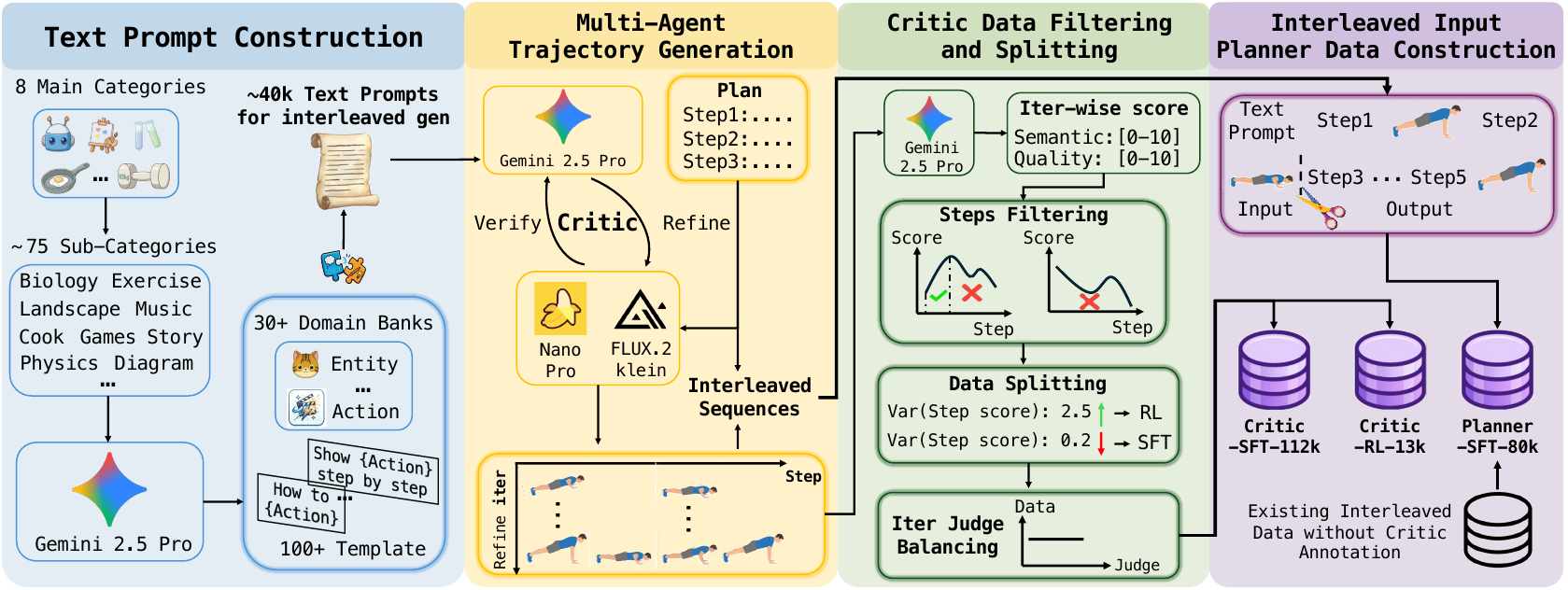}
    \vspace{-4mm}
    \caption{\small\textbf{Illustration of Our Data Construction Pipeline}.}
    \label{fig:data_con}
    \vspace{-2mm}
\end{figure}

\subsection{Dataset Construction Pipeline}
\label{subsec:data}

High-quality training data is essential for developing agents capable of long-horizon planning and step-wise correction. However, aligned pairs of interleaved instructions, intermediate visual states, and critic judgement, refinements, thinking process do not naturally exist. To address this, as shown in Fig~\ref{fig:data_con}, we construct a dedicated data pipeline comprising four main stages.

\textbf{Text Prompt Construction.} We curate a comprehensive set of text prompts that covers primary interleaved generation tasks (visual narrative, guidance, and embodied manipulation). To ensure dataset diversity, we propose a systematic, top-down generation pipeline. We initiate this process by defining 8 main categories spanning broad domains, including robotics, visual storytelling, art, workflows, daily life, science, and professional skills. These main categories are further divided into approximately 75 fine-grained sub-categories, such as biology, cooking, and physics. We then prompt Gemini 2.5 Pro~\cite{gemini25pro} to expand these sub-categories into more than 30 domain-specific vocabulary banks, extracting key entities and actions. Finally, we populate over 100 predefined instructional templates (e.g., ``How to \{Action\}'', ``Show \{Action\} step by step'') with elements from these domain banks. This procedural generation approach ultimately yields roughly 40,000 diverse text prompts tailored for interleaved generation.

\textbf{Multi-Agent Trajectory Generation.} Given the collected prompts, we employ advanced proprietary models, Gemini 2.5 Pro~\cite{gemini25pro} and Nano Banana Pro~\cite{nanopro}, to generate agentic trajectories. For each task, the Planner agent first generates a global step-by-step instruction sequence. Then, an image generator (\textit{i.e.}, since the trajectory data generated by Nano Banana Pro is of exceptionally high quality, we introduce FLUX.2-klein-9B to balance the dataset, thereby preventing the Critic from becoming biased.) executes these instructions step-by-step. At each step, the Critic agent evaluates the generated image, compares it against the Planner's original instruction, and produces a critique. If the image deviates from the instruction, the Critic refines the prompt for immediate regeneration. This iterative process yields complete trajectories containing global plans, intermediate images, critiques, and refined prompts.

\textbf{Critic Data Filtering and Splitting.} To ensure the quality of the synthesized trajectories, we apply a rigorous filtering pipeline that eliminates samples with severe logical inconsistencies or poor visual quality. Note that \textit{this filtering process is exclusively applied to curate the training data for the Critic}, while the training data for the Planner remains unfiltered. Since optimizing an entire interleaved trajectory (\textit{One trajectory maybe consist of 25 generator calls}) via RL is computationally prohibitive and unstable, we first decompose the generated trajectories into independent step-wise data. This decomposition enables a single-iteration optimization approach (as detailed in Sec. 3.4). We then employ Gemini 2.5 Pro~\cite{gemini25pro} with an adapted system prompt from VIEScore~\cite{ku2024viescore} to evaluate every refinement iteration within each step, assigning scores from 0 to 10 for both semantic alignment and visual quality. Based on these iteration-level scores, we process the step-wise data through the following three stages.

\textbf{1) Steps Filtering.} We analyze the progression of the Gemini 2.5 Pro scores across the refinement iterations within each step. As illustrated in the scoring curves, we discard steps that exhibit negative refinement trends, score degradation, or persistent low quality. Only the steps demonstrating successful refinement, characterized by an upward or stable high-score trajectory, are retained for subsequent processing.
\textbf{2) SFT-RL Data Splitting.} To construct tailored datasets for SFT and RL, we compute the variance of the iteration scores within each valid step. Steps with a high score variance indicate a dynamic refinement process with substantial quality shifts, making them ideal for RL optimization. Conversely, steps with low variance represent stable and high-quality generation, which are better suited for the SFT dataset. We partition the data accordingly and maintain an empirical sample ratio of 2:1 between the SFT and RL subsets.
\textbf{3) Iter-wise Judgment Distribution Balancing.} The Critic includes an objective to predict the binary judgment of a given iteration. Training on the natural, heavily skewed data leads to biased estimations. To address this, we balance the iteration-wise data by resampling the samples as shown in Fig~\ref{fig:data_con}. Ultimately, this process yields two high-quality datasets: \texttt{Interleave-Critic-SFT-112k} for SFT, and \texttt{Interleave-Critic-RL-13k} for RL.

\textbf{Interleaved Input Planner Data Construction.} Since our initial instructions consist solely of pure text prompts, the resulting dataset naturally lacks the multimodal interleaved context required to train the Planner. To address this limitation, we adopt two strategies to construct interleaved input-output pairs. First, we generate self-synthesized interleaved trajectories by interleaving the previously generated textual plans with their corresponding final image outputs at each step. To formulate training pairs, we randomly select a step to truncate this sequence. The sequence preceding the truncation point acts as the interleaved multimodal input, while the subsequent text plan is assigned as the target output. Second, we leverage existing open-source interleaved datasets~\cite{chen2024comm}. Although these datasets lack the fine-grained annotations necessary for training the Critic, their natural text-image structures are perfectly suited for the Planner. Consequently, the final training corpus for the Planner is composed of both the self-synthesized truncated sequences and the external unannotated interleaved data. Ultimately, this process yields \texttt{Interleave-Planner-SFT-80k}

\subsection{Training Scheme}
\label{subsec:training}

Based on the constructed datasets, we train the InterleaveThinker framework through a two-stage pipeline: SFT for multi-agent format cold-start, followed by RL to reinforce the Critic's correction capabilities using GRPO.

\textbf{Planner-SFT.} The Planner is initialized with Qwen3-VL-8B-Instruct~\cite{qwen3vl} and fine-tuned using the \texttt{Interleave-Planner-SFT-80k} dataset. Details regarding the system prompt and SFT format can be found in Appendix~\ref{eval_prompt}. The SFT training equips the model with the ability to break down a complex user request into a coherent, global sequence of text-image instructions upfront, thereby bypassing the visual over-reliance problem. Note that we did not apply RL to the Planner. Because our trajectories can involve over 25 rounds of generator tool calls, the reward signals become highly sparse, making RL optimization highly unstable. Furthermore, since SFT alone already achieves strong performance, RL was deemed unnecessary.

\textbf{Critic-SFT.} Critic is initialized with Qwen3-VL-8B-Instruct~\cite{qwen3vl}, SFT teaches the model the basic format of evaluation: observing the current visual state, identifying deviations from the planned instruction, and formulating a refined prompt for the generator. We show that format below and the system prompt is shown in Appendix~\ref{eval_prompt}.
$$
\texttt{<think></think><answer>[Judgment][Refined Prompt]</answer>}
$$
\textbf{Dual-Reward Strategy for Efficient Critic RL.} 
A unique challenge in applying RL to interleaved generation is the extreme length of the generation trajectories. A single interleaved task may require over 25 generator calls. Optimizing the entire trajectory end-to-end using standard RL algorithms introduces prohibitive computational costs and severe credit assignment issues. 

To resolve this, we propose a single-step RL formulation guided by a dual-reward strategy to effectively simulate full-trajectory optimization. Since our decoupled Planner generates all step-by-step instructions upfront, the generation process naturally breaks down into independent stages. Within each step, the Critic evaluates the output and iteratively generates refinement prompts until a satisfactory quality threshold is met, allowing the system to seamlessly advance to the next pre-planned instruction. Consequently, ensuring the success of each local iteration guarantees the overall success of the global trajectory. The \textbf{Accuracy Reward ($R_{acc}$)} measures the Critic's ability to accurately judge the current generation by penalizing the difference between its predicted one and the ground truth $J_i$, ensuring reliable threshold identification. The formulation is as:
\begin{equation}
    R_{acc} = -|\texttt{Critic}\left(I_{i-1}, I_i^t, p_i, r_i^t\right) - J_i|.
\end{equation}
Meanwhile, the \textbf{Step-wise Reward ($R_{step}$)} evaluates the effectiveness of the Critic's interventions when an output falls below the threshold. It is computed as the score difference between the newly iteration result $I_i^{t+1}$ and the original $I_i^t$, the formulation is as
\begin{equation}
    R_{step} = \texttt{Gemini}\left(I_{i-1}, I_i^{t+1}, p_i, r_i^{t+1}\right) - \texttt{Gemini}\left(I_{i-1}, I_i^t, p_i, r_i^t\right),
\end{equation}
where a positive delta indicates that the refinement prompt successfully improved the output, directly rewarding actionable and effective critiques. Note that we use expert Gemini 2.5 Pro to score the result to ensure the accuracy and consistency with binary judgment.
The final reward for a single correction step is computed as a weighted combination of both signals and the format reward $R_{format}$:
\begin{equation}
    R = 0.5 * R_{format} + 0.5 * (\alpha R_{acc} + (1 - \alpha) R_{step})
\end{equation}
where $\alpha$ is a balancing hyperparameter and set to 0.2 by default. By normalizing these rewards within a sampled group, we compute the advantages and update the Critic's policy using the GRPO objective. For the implementation details about GRPO, please refer to~\cite{guo2025deepseek}.

\setlength{\fboxrule}{1pt}
\begin{table}[t]
\scriptsize 
  \caption{\small \textbf{Comparison on UEval~\cite{li2026ueval}.} We evaluate open-source and proprietary frontier models on 8 tasks in UEval. \textbf{Bold} indicates the best result among each group.}
  \label{tab:ueval}
  \centering
{\setlength{\tabcolsep}{5pt}  
  \begin{tabular}{l ccccccccc}
    \toprule[0.1em]
    \textbf{Models} & \textbf{Space} & \textbf{Textbook} & \textbf{Diagram} & \textbf{Paper} & \textbf{Art} & \textbf{Life} & \textbf{Tech} & \textbf{Exercise} & \textbf{Avg} \\
    \midrule[0.1em]
    \rowcolor{gray!10}
    \multicolumn{10}{c}{\textbf{\textit{Reference}}} \\
    \midrule[0.1em]
    \textcolor{gray}{Reference} & \textcolor{gray}{96.2} & \textcolor{gray}{94.4} & \textcolor{gray}{93.1} & \textcolor{gray}{96.2} & \textcolor{gray}{90.6} & \textcolor{gray}{87.7} & \textcolor{gray}{90.6} & \textcolor{gray}{89.2} & \textcolor{gray}{92.2} \\
    \midrule[0.1em]
    \rowcolor{gray!10}
    \multicolumn{10}{c}{\textbf{\textit{Proprietary Frontier Models}}} \\
    \midrule[0.1em]
    Gemini-2.0-Flash~\cite{gemini2flash} & 65.2 & 55.2 & 47.6 & 45.8 & 70.4 & 58.0 & 50.2 & 48.0 & 55.1 \\
    GPT-5-Instant~\cite{gpt5} & 77.3 & 77.9 & 62.3 & 55.1 & 71.2 & 69.7 & 50.7 & 57.6 & 65.2 \\
    GPT-5-Thinking~\cite{gpt5} & 84.0 & 78.0 & 67.8 & 51.9 & 67.8 & 63.8 & 57.0 & 61.4 & 66.4 \\
    Nano Banana~\cite{nanobanana} & 78.0 & 74.0 & 66.4 & 71.6 & 66.6 & 63.0 & 58.2 & 50.0 & 66.0 \\
    Nano Banana Pro~\cite{nanopro} & 79.4 & 89.6 & 75.9 & 81.3 & 84.3 & 73.5 & 60.8 & 63.9 & 76.1 \\
    \midrule[0.1em]
    \rowcolor{gray!10}
    \multicolumn{10}{c}{\textbf{\textit{Open-Sourced Models}}} \\
    \midrule[0.1em]
    Janus-Pro~\cite{januspro} & 21.0 & 31.0 & 37.4 & 15.2 & 26.4 & 23.0 & 17.6 & 11.5 & 22.9 \\
    Show-o2~\cite{xie2025showo2} & 25.4 & 33.1 & 33.2 & 17.4 & 25.6 & 15.6 & 17.4 & 13.1 & 22.6 \\
    MMaDA~\cite{yang2025mmada} & 10.8 & 20.0 & 14.2 & 13.3 & 15.7 & 15.8 & 12.4 & 12.6 & 14.4 \\ 
    BAGEL~\cite{bagel} & 29.8 & 42.5 & 37.2 & 20.0 & 39.0 & 33.6 & 24.8 & 21.4 & 31.0 \\
    Emu3.5~\cite{cui2025emu3_5} & 59.1 & 57.4 & 41.1 & 31.6 & 59.3 & 62.0 & 37.0 & 45.4 & 49.1 \\
    \midrule[0.1em]
    InterleaveThinker+FLUX.2-klein-9B & 62.1 & 92.0 & 82.1 & 75.1 & 71.0 & 54.6 & 36.6 & 43.8 & \textbf{66.3} \\
    InterleaveThinker+Qwen-Image-Edit & 65.8 & 90.5 & 84.2 & 77.9 & 70.4 & 55.7 & 36.3 & 44.2 & \textbf{67.2} \\
    \bottomrule[0.1em]
  \end{tabular}
  }
  \vspace{-1em}
\end{table}

\section{Experiments}
\label{sec:exp}
\subsection{Experimental Setup}
\label{subsec:imp_detail}
\textbf{Implementation Details.} Both the Planner and the Critic are initialized from the Qwen3-VL-8B-Instruct model. In the SFT stage, the Planner and the Critic are both trained for two epochs, using a learning rate of $2 \times 10^{-5}$ and a batch size of 32. Then, the Critic is trained for one epoch of RL. For the RL stage, we set the learning rate to $2 \times 10^{-6}$, the global batch size to 16, the rollout number ($N$) to 8, and apply a KL divergence penalty with a coefficient of $1 \times 10^{-3}$. Throughout the training, the maximum image resolution is capped at $1024 \times 1024$. The entire pipeline takes approximately 50 hours on eight H800 GPUs. During inference, we integrate InterleaveThinker with three distinct models to evaluate different aspects of our approach and set the maximum refinement iteration $T_{max}$ for each step to 5. We use FLUX.2-klein-9B~\cite{flux2} for in-domain evaluation and Qwen-Image-Edit-2511~\cite{qwenimage} to assess generalization capabilities. 

\textbf{Benchmarks.} To systematically evaluate the capabilities of our multi-agent InterleaveThinker, we test it on two interleaved benchmarks: UEval~\cite{li2026ueval} and CoMM~\cite{chen2024comm} (Tasks 3 and 4). Specifically, UEval assesses text-to-interleaved output generation, while task3 of CoMM measures interleaved input-output performance. Furthermore, we validate our method on reasoning-based benchmarks, utilizing WISE~\cite{niu2025wise} for image generation and RISE~\cite{rise} for image editing.

\begin{figure}
    \centering
    \includegraphics[width=0.7\linewidth]{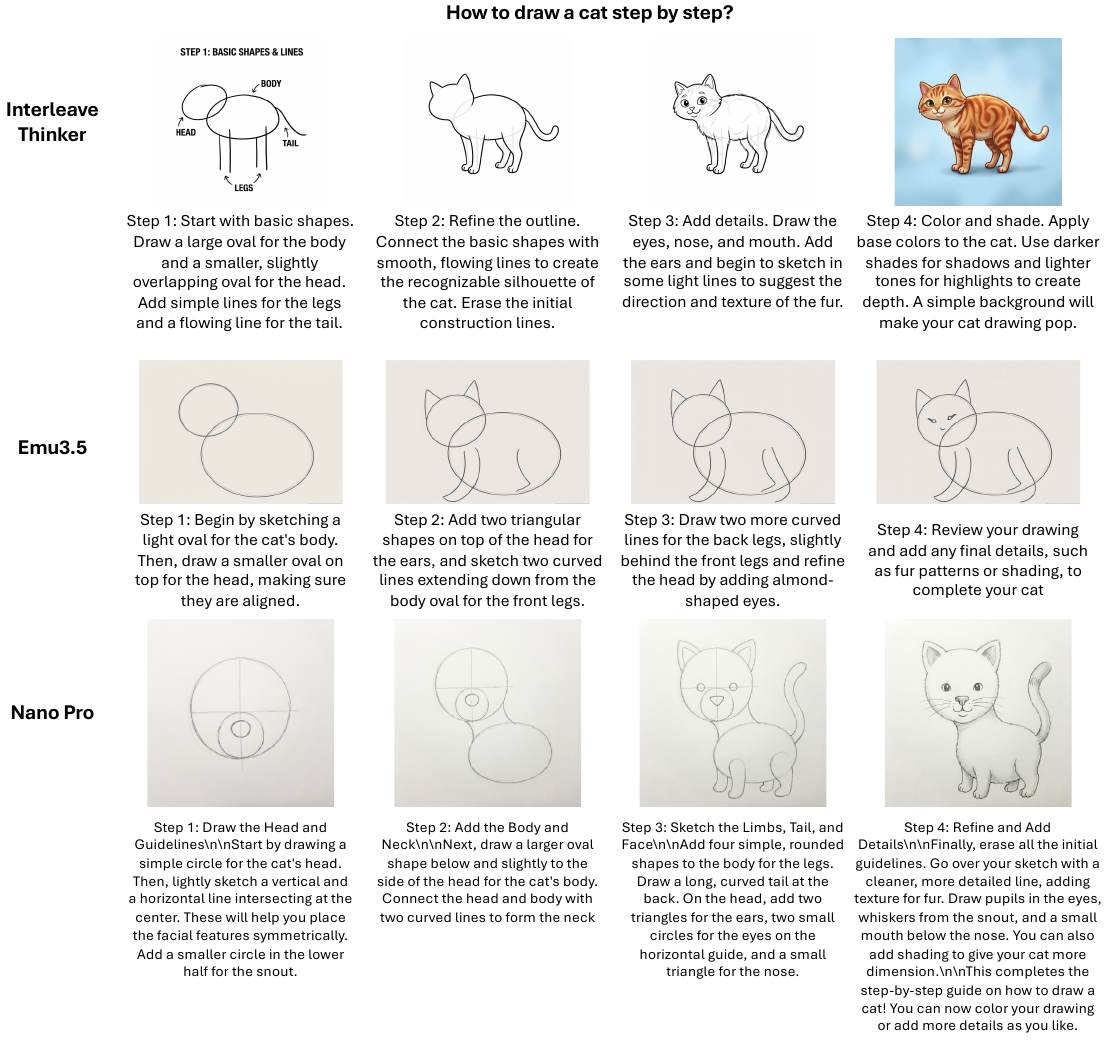}
    \vspace{-3mm}
    \caption{\small\textbf{Comparison with Emu3.5 and Nano Banana Pro in pure-text input interleaved generation}.}
    \label{fig:per1}
    \vspace{-4mm}
\end{figure}

\begin{figure}
    \centering
    \includegraphics[width=0.8\linewidth]{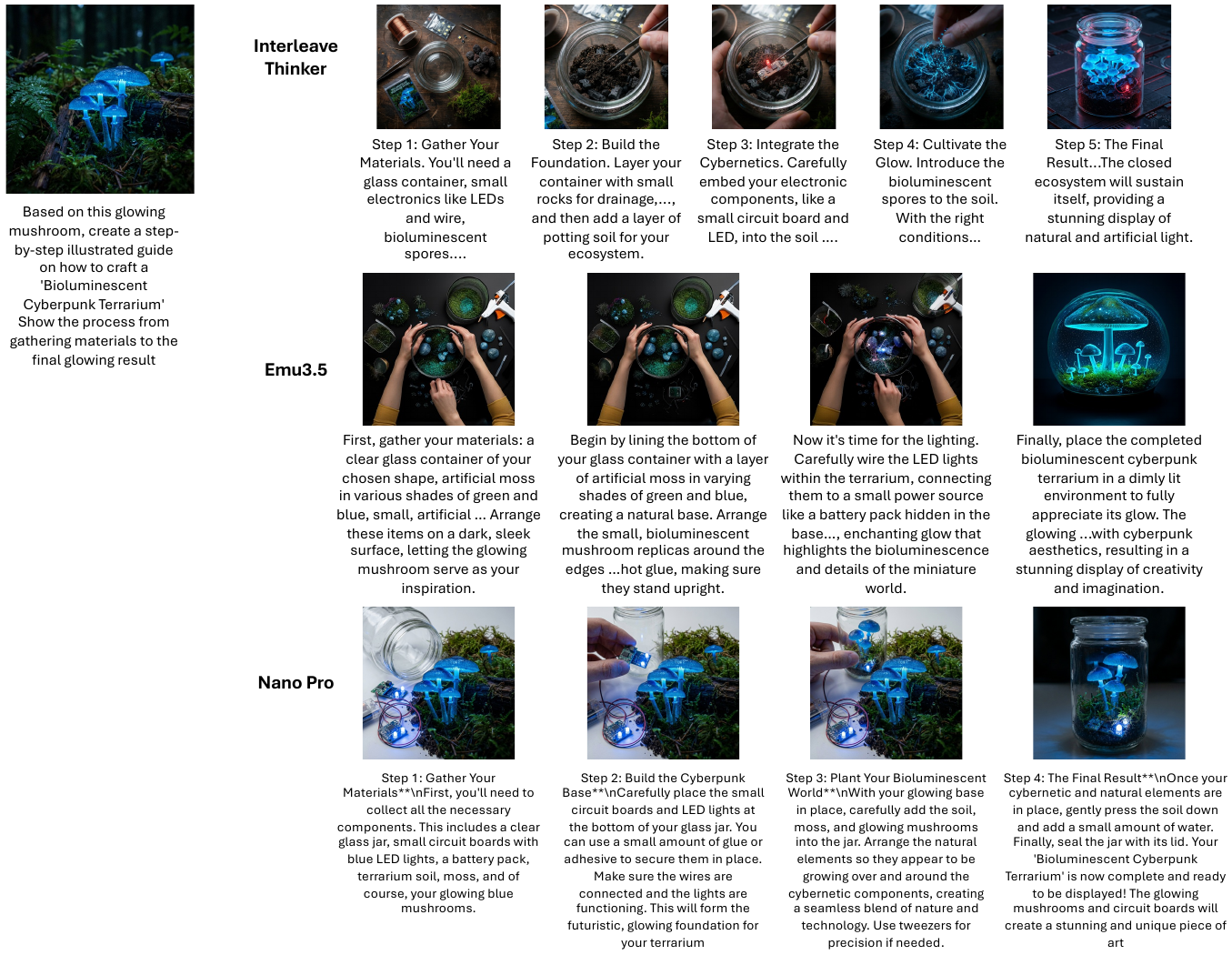}
    \vspace{-3mm}
    \caption{\small\textbf{Comparison with Emu3.5 and Nano Banana Pro in multi-modal input interleaved generation}.}
    \label{fig:per2}
    % \vspace{-2mm}
\end{figure}

\begin{table}[t]
\small 
  \caption{\textbf{Comparison on CoMM~\cite{chen2024comm}.} Sty. and Enti. denotes the style and entity consistency among generated images. Tren. denotes the trend alignment betwen image and text squence. Comp. denotes the completeness, ImgQ is the image quality. IRS means text-image alignment score. x/x reflects the model's performance on interleaved (Task 3) and pure-text (Task 4) inputs.}
  \label{tab:comm_comparison}
  \centering
{\setlength{\tabcolsep}{5pt}  
  \begin{tabular}{l cccccc}
    \toprule[0.1em]
    Model & Sty. & Enti. & Tren. & Comp. & ImgQ & IRS \\
    \midrule[0.1em]
    MiniGPT-5~\cite{zheng2023minigpt}  & 5.6 / 5.7 & 5.2 / 5.2  & 5.2 / 5.3 & 6.3 / 5.8 & 6.4 / 6.2 & 2.6 / 2.7 \\
    SEED-LLaMA~\cite{seedllama} & 6.3 / 7.6 & 5.8 / 6.8 & 5.7 / 6.2 & 6.3 / 5.1 & 6.6 / 6.4 & 2.9 / 1.5 \\
    Emu2~\cite{emu2}     & 8.2 / 8.4 & 8.0 / 7.6 & 8.0 / 7.6 & 8.5 / 7.5 & 8.6 / 7.6 & 2.4 / 2.0 \\
    DuoGen~\cite{duogen}   & - / 9.2 & - / 9.2 & - / 9.2 & - / 9.7 & - / 9.5 & - / 7.8 \\
    \midrule[0.1em]
    InterleaveThinker+FLUX.2-klein-9B & \textbf{9.3} / \textbf{9.6} & \textbf{9.2} / 9.6 & \textbf{9.1} / 9.5 & 9.1 / 9.6 & \textbf{9.7} / \textbf{9.8} & 5.2 / \textbf{8.2}\\
    InterleaveThinker+Qwen-Image-Edit & 9.2 / \textbf{9.6} & 9.1 / \textbf{9.7} & 9.0 / \textbf{9.6} & \textbf{9.2} / \textbf{9.8} & \textbf{9.7} / \textbf{9.8} & \textbf{5.5} / \textbf{8.4}\\
    \bottomrule[0.1em]
  \end{tabular}
  }
  \vspace{-1em}
\end{table}

\begin{table}[t]
\small 
  \caption{\textbf{Comparison on WISE~\cite{niu2025wise}.} \textbf{Bold} indicates the best result among each group.}
  \label{tab:wise_benchmark_comparison}
  \centering
{
\setlength{\tabcolsep}{3pt}  
  \begin{tabular}{l ccccccc}
    \toprule[0.1em]
    \textbf{Model} & \textbf{Cultural} & \textbf{Time} & \textbf{Space} & \textbf{Biology} & \textbf{Physics} & \textbf{Chemistry} & \textbf{Overall} \\
    \midrule[0.1em]
    \rowcolor{gray!10}
    \multicolumn{8}{c}{\textbf{\textit{Proprietary Frontier Models}}} \\
    \midrule[0.1em]
    GPT-Image-1~\cite{gptimage}         & 0.81 & 0.71 & 0.89 & 0.83 & 0.79 & 0.74 & 0.80 \\
    Nano Banana Pro~\cite{nanopro}     & 0.89 & 0.80 & 0.89 & 0.88 & 0.86 & 0.85 & 0.87 \\
    \midrule[0.1em]
    \rowcolor{gray!10}
    \multicolumn{8}{c}{\textbf{\textit{Open-Sourced Models}}} \\
    \midrule[0.1em]
    SD-3.5-large~\cite{sd35l}        & 0.44 & 0.50 & 0.58 & 0.44 & 0.52 & 0.31 & 0.46 \\
    FLUX.1-dev~\cite{flux2024}          & 0.48 & 0.58 & 0.62 & 0.42 & 0.51 & 0.35 & 0.50 \\
    Hunyuan-Image-3.0~\cite{cao2025hunyuanimage} & 0.57 & 0.58 & 0.75 & 0.58 & 0.71 & 0.47 & 0.61 \\
    Qwen-Image~\cite{qwenimage}   & 0.62 & 0.63 & 0.77 & 0.57 & 0.75 & 0.40 & 0.62 \\
    LongCat-Image~\cite{longcatimage}  & 0.66 & 0.61 & 0.72 & 0.66 & 0.72 & 0.49 & 0.65 \\
    BAGEL~\cite{bagel}  & 0.76 & 0.69 & 0.75 & 0.65 & 0.75 & 0.58 & 0.72 \\
    \midrule[0.1em]
    FLUX.2-klein-9B~\cite{flux2}  & 0.44 & 0.60 & 0.67 & 0.32 & 0.50 & 0.27 & 0.47 \\
    \quad\textbf{+InterleaveThinker (Ours)} & 0.72 & 0.70 & 0.82 & 0.72 & 0.78 & 0.69 & \textbf{0.73}  \\
    Qwen-Image-Edit-2511~\cite{qwenimage} & 0.60 & 0.60 & 0.76 & 0.52 & 0.66 & 0.39 & 0.60 \\
    \quad\textbf{+InterleaveThinker (Ours)} & 0.74 & 0.67 & 0.83 & 0.72 & 0.76 & 0.56 & \textbf{0.72}  \\
    \bottomrule[0.1em]
  \end{tabular}
  }
  % \vspace{-4mm}
  \vspace{-1em}
\end{table}

\begin{table}[t]
\small 
  \caption{\textbf{Comparison on RISE-Bench~\cite{rise}.}}
  \label{tab:rise}
  \centering
{
\setlength{\tabcolsep}{7pt}  
  \begin{tabular}{l ccccc}
    \toprule[0.1em]
    \textbf{Model} & \textbf{Temporal} & \textbf{Causal} & \textbf{Spatial} & \textbf{Logical} & \textbf{Overall} \\
    \midrule[0.1em]
    \rowcolor{gray!10}
    \multicolumn{6}{c}{\textbf{\textit{Proprietary Models}}} \\
    \midrule[0.1em]
    Seedream-4.0~\cite{seedream4} & 12.9 & 12.2 & 11.0 & 7.1 & 10.8 \\
    GPT-Image-1~\cite{gptimage} & 34.1 & 32.2 & 37.0 & 10.6 & 28.9 \\
    Nano Banana~\cite{nanobanana} & 25.9 & 47.8 & 37.0 & 18.8 & 32.8 \\
    Nano Banana Pro~\cite{nanopro} & 41.2 & 61.1 & 48.0 & 37.6 & 47.2 \\
    \midrule[0.1em]
    \rowcolor{gray!10}
    \multicolumn{6}{c}{\textbf{\textit{Open-source Models}}} \\
    \midrule[0.1em]
    Step1X-Edit~\cite{step1x} & 0.0 & 2.2 & 2.0 & 3.5 & 1.9 \\
    Ovis-U1~\cite{wang2025ovis} & 1.2 & 3.3 & 4.0 & 2.4 & 2.8 \\
    FLUX.1-Kontext-Dev~\cite{flux-kontext} & 2.3 & 5.5 & 13.0 & 1.2 & 5.8 \\
    BAGEL~\cite{bagel} & 2.4 & 5.6 & 14.0 & 1.2 & 6.1 \\
    BAGEL (w/ CoT)~\cite{bagel}  & 5.9 & 17.8 & 21.0 & 1.2 & 11.9 \\
    Qwen-Image-Edit~\cite{qwenimage} & 4.7 & 10.0 & 17.0 & 2.4 & 8.9 \\
    \midrule[0.1em]
    FLUX.2-klein-9B~\cite{flux2}  & 7.1 & 13.3 & 24.0 & 7.1 & 13.3 \\
    \quad\textbf{+InterleaveThinker (Ours)} & 36.5 & 33.3 & 34.0 & 10.6 & \textbf{28.9}\\
    Qwen-Image-Edit-2511~\cite{qwenimage} & 21.2	& 18.9 & 31.0 & 4.7 & 19.4\\
    \quad\textbf{+InterleaveThinker (Ours)} & 27.1 & 38.9 & 39.0 & 12.9 & \textbf{30.0} \\
    \bottomrule[0.1em]
  \end{tabular}
  }
  \vspace{-2mm}
\end{table}

\subsection{Main Results.}
\vspace{-1mm}
\textbf{Results on UEval.} As summarized in Table~\ref{tab:ueval}, our multi-agent pipeline significantly outperforms existing open-source UMMs and achieves performance comparable to the highly capable Nano Banana. More importantly, the further performance gains observed when integrating with Qwen-Image-Edit demonstrate that InterleaveThinker is a model-agnostic and highly generalizable framework.

\textbf{Results on CoMM.} As shown in Table~\ref{tab:comm_comparison}, InterleaveThinker surpasses all existing methods even when solely integrated with the 4-step FLUX.2-klein. Furthermore, applying our framework to stronger models like Qwen-Image-Edit-2511 further pushes the performance boundaries on this benchmark.

\textbf{Results on WISE.} It is important to note that neither our Planner nor our Critic was explicitly trained on reasoning-based image generation tasks. Remarkably, the results in Table~\ref{tab:wise_benchmark_comparison} show that our method significantly improves upon the base models. 
This demonstrates that our multi-agent plan-generate-critic framework is also highly beneficial for reasoning-based image generation tasks.

\textbf{Results on RISE.} The performance on the reasoning-based image editing task mirrors the success observed on WISE. 
As shown in Table~\ref{tab:rise}, our approach significantly improves the base models.

\textbf{Visualization.} We further provide qualitative visual comparisons in Fig~\ref{fig:per1} and Fig~\ref{fig:per2}. 
InterleaveThinker effectively mitigates the problems of visual over-reliance and step-wise error accumulation, while simultaneously maintaining high textual fidelity and superior image quality.

\subsection{Ablation Study}
\label{subsec:ablation}

We conduct extensive ablation studies on the UEval benchmark and use FLUX.2-klein-9B as the default image generator. For reference, we also report the upper-bound performance achieved by two proprietary oracle models (Gemini-2.5-Pro and GPT-4.1). The results are shown in Table~\ref{tab:ablation}.

\textbf{Effectiveness of Multi-Agent workflow.} 
The raw FLUX.2-klein-9B generator alone fails entirely at interleaved generation due to output limitation. 
To establish a zero-shot multi-agent baseline, we deploy the Qwen3-VL-8B-Instruct model as both the Planner and the Critic. When we introduce the \textit{Planner-SFT} module (while keeping the Critic as the zero-shot Qwen3-VL-8B-Instruct), we observe a massive surge in the Text score from $33.5$ to $58.5$. 
Subsequently, upgrading the pipeline to \textit{Full-SFT} (where both the Planner and Critic are fine-tuned) further boosts the Image quality. This confirms that the Critic-SFT successfully identify visual deviations and provide actionable corrections that the zero-shot model cannot formulate.

\begin{wraptable}{r}{0.5\textwidth}
\centering
\caption{\textbf{Ablation Study on UEval.}}
\vspace{-2mm}
\resizebox{0.5\textwidth}{!}{
\begin{tabular}{l@{\hspace{1em}}c@{\hspace{1em}}c@{\hspace{1em}}cccc}
    \toprule
    \textbf{Model} & \textbf{Text} & \textbf{Image} & \textbf{Avg} \\
    \midrule[0.12em]
    FLUX.2-klein-9B          & 0 & 36.4 & 18.2 \\
    \midrule[0.1em]
    + Gemini-2.5-pro (oracle)  & 74.8 & 79.9 & 77.4 \\
    + GPT 4.1   &  63.2 & 71.8 & 67.5 \\
    \midrule[0.1em]
    + Qwen3-VL-8B (Baseline)  & 33.5 & 62.6 & 48.1 \\
    + Planner-SFT  & 58.5 & 61.8 & 60.5  \\
    % + Critic-SFT           &  \\
    + Full-SFT & 58.6 & 70.4 & 64.5  \\
    + RL w/o step reward & 58.2 & 72.2 & 65.2 \\
    + RL w/o acc reward  & 58.4 & 71.7 & 65.1 \\
    + Full-RL & \textbf{58.6} & \textbf{74.0} & \textbf{66.3} \\
    \midrule[0.1em]
    One-Agent & 45.2 & 63.7 & 54.5\\
    \midrule[0.1em]
    Unfiltered data & 58.2 & 67.3 & 62.8\\
    \midrule[0.1em]
    $T_{max}=1$ & 58.5 & 61.8 & 60.2\\
    $T_{max}=3$ & 58.6 & 72.0 & 65.3\\
    $T_{max}=5$ & 58.6 & 74.0 & 66.3\\
\bottomrule
\end{tabular}}
\label{tab:ablation}
\vspace{-6mm}
\end{wraptable}

\textbf{Impact of the Dual-Reward RL Scheme.} 
We ablate the reward signals used in the RL stage. Removing the Step-wise Reward ($R_{step}$) decreases the average score, as the Critic fails to optimize the refined prompts effectively. Conversely, removing the Accuracy Reward ($R_{acc}$) drops the score as it leads to inaccurate score evaluation. Ultimately, combining both rewards yields the best result.

\textbf{Multi-Agent vs. One.} 
To further validate the issue of visual over-reliance in single VLM, we integrated the planner's capabilities into the critic, allowing the model to simultaneously plan the next step and evaluate the previous one. The results indicate that this paradigm severely degrades model performance when the image generator is frozen, corroborating our claim in the introduction.

\textbf{Importance of Critic Data Filtering.} 
In our dataset construction pipeline (Sec.~\ref{subsec:data}), we introduced step filtering and iteration-wise judgment distribution balancing. We train an ablation variant of the Critic using the unfiltered data. This Critic tends to collapse into trivial constant predictions (e.g., frequently output \texttt{True} regardless of the actual image quality), leading to performance drop.

\textbf{Influence of Maximum Refinement Iterations.} 
InterleaveThinker's closed-loop refinement relies on the maximum iteration count $T_{max}$. Increasing $T_{max}$ consistently improves performance over the single-pass baseline ($T_{max}=1$), demonstrating the Critic's effectiveness.

\section{Conclusion and Limitations}
\label{sec:conclu}
In this work, we identify that existing multimodal models struggle with long-horizon interleaved generation due to visual over-reliance and step-wise error accumulation. We attribute this to the entangled planning and visual evaluation within a single model, and propose a decoupled multi-agent framework, InterleaveThinker, to address it. InterleaveThinker consists of a Planner that predicts global instructions upfront to bypass visual interference, and a Critic agent that performs step-wise evaluation and prompt refinement. To overcome the computational bottleneck of long-trajectory RL, we further introduce a dual-reward strategy that enables efficient single-step RL on the Critic to guide the entire generation sequence. Extensive experiments show that InterleaveThinker endows off-the-shelf image generators with strong interleaved generation capabilities, matching proprietary models while surprisingly boosting complex reasoning performance.

\textbf{Limitations.} Although adaptable to any image generator, our framework's capacity is constrained by the base model's generative prior. Consequently, it cannot generate concepts that were not included in the base generator's training corpus. We further show the bad case about this in Fig~\ref{fig:bad_case} in Appendix.

{
    \small
    \bibliographystyle{unsrt}
    \bibliography{conference}
}

\clearpage

\appendix

\section{System Prompt}
\label{eval_prompt}

\begin{tcolorbox}[
    colback=lightgray!10,
    colframe=black,
    title={\textbf{Planner Pure-Text System Prompt}},
    breakable
]
\vspace{0.5em}
\begin{Verbatim}[breaklines=true, breaksymbol={}, fontsize=\tiny]
# Task Planner, Orchestrator, and Prompt Engineer System

You are an expert **Task Planner, Orchestrator, and Prompt Engineer**.
Your goal is to analyze a user's request, generate a structured execution plan, and optimize EVERY step's instruction into a highly effective Text-to-Image (T2I) prompt or Image Editing instruction.

## Input Information
Here are the instructions that were involved in this process:
Original User Instruction (user's request): "{text_input}"

## Execution Plan Instructions
1. **Dynamic Step Count (Image Operations Only)**: Determine the necessary number of steps. Every step in your execution plan MUST represent an actual image generation or image editing action. **DO NOT** create separate steps solely for generating text, captions, or summaries. 
2. **Complete & Polished Output**: Always aim for a fully realized final product. For visual or creative tasks, the final step MUST result in a fully colored, detailed, and polished output. Do not stop at a draft, outline, or uncolored sketch unless the user explicitly requests it.
3. **Text Generation & Auxiliary Text Rule**: 
   - If the user specifically asks to render or draw text *inside* the image, include this requirement within the `instruction` field.
   - If the user explicitly asks for a *separate* text response (e.g., a caption, summary, explanation, or knowledge grounding) to accompany the image, generate this text and place it in the `auxiliary_text` field of the corresponding image generation step. 
   - If the user does not explicitly request any separate text or caption, you MUST set `auxiliary_text` to `null`.

## Optimize Prompt Instructions
1. **Prompt Optimization for All Steps**: Convert the `instruction` of EVERY step into a highly effective prompt in the `prompt` field.
   - **Step 1 (Generation)**: Create a highly detailed T2I prompt representing the foundational stage. Focus *only* on the Step 1 instruction. Do NOT hallucinate unmentioned details or future elements.
   - **Subsequent Steps (Editing)**: Create clear, actionable image editing instructions (e.g., "add a red hat", "change the background to a cyberpunk city") based on the current step's goal.
2. **CRITICAL**: The `prompt` field MUST contain ONLY the pure text prompt or editing instruction. DO NOT include meta-text, prefixes (such as "Step 1:", "Prompt:", "Edit:"), or conversational filler. It must be directly usable by the generation/editing API.

## Output
The output consists of two parts:
1. A Statement - Analysis process and reasoning;
2. A JSON — Planing each step and rewrite the instruction to prompt suitable for generation/editing.

Here is a output example

<think>
Part 1: Planning analysis explaining the execution plan. Part 2: Analysis of how the instructions were translated into visual keywords for the T2I prompt and editing instructions.
</think>

<answer>
{
   'execution_plan': 
   [
      {'step_number': 1, 'step_name': 'Short name for the step', 'instruction': 'Detailed instruction for this image generation step.', 'prompt': "The optimized, pure T2I prompt suitable for the image generation model. (No 'Step 1:' prefix)", 'auxiliary_text': 'The required caption, summary, or text explanation. Output null if no separate text is explicitly requested.'},
      {'step_number': 2, 'step_name': 'Short name for the step', 'instruction': 'Detailed instruction for this image editing step.', 'prompt': "The optimized, pure instruction suitable for the image editing model. (No 'Step 2:' prefix)", 'auxiliary_text': None}
   ]
}
</answer>
\end{Verbatim}
\end{tcolorbox}

\begin{tcolorbox}[
    colback=lightgray!10,
    colframe=black,
    title={\textbf{Planner Interleaved System Prompt}},
    breakable
]
\vspace{0.5em}
\begin{Verbatim}[breaklines=true, breaksymbol={}, fontsize=\tiny]
You are an expert **Multimodal Sequence Planner and Orchestrator**.
Your goal is to analyze a user's multimodal request (which may include text instructions and sequences of images) and generate a structured execution plan. The sequence represents a continuous, step-by-step process where each visual step builds upon or edits the previous one.

## Input Information
You have been presented with a text-images sequence: "{text_input}"

### Instructions
1. **Task Identification & Modality Routing**: Carefully analyze the input to determine the task type.
   - **Task A (General Text Response / Problem Solving / Image-to-Text)**: If the user provides a complete sequence of images and asks for text responses for each step (e.g., describing the images, solving a problem, explaining a process, or answering questions), you must write your complete response entirely within the `auxiliary_text` field. You MUST set BOTH the `instruction` and `prompt` fields to `null` for these steps.
   - **Task B (Sequence Continuation / Sequential Editing)**: If the user provides a partial sequence and asks to predict/generate the remaining steps, you must generate both the text instruction and the editing prompt. The `prompt` field must contain an optimized instruction specifically tailored for an **image editing model** to modify the previous step's image into the new state.
2. **Strict Step Count & NO Prefix Rule**: 
   - **Step Count**: Determine the logical number of steps. **CRITICAL**: If the user's input explicitly specifies the number of steps required, you MUST strictly output exactly that number of steps to fulfill the requirement. If continuing a sequence (Task B), your `step_number` MUST start exactly from where the user's input left off.
   - **NO Prefixes**: BOTH the `instruction` and `prompt` fields MUST NOT contain any step prefixes, numbers, or bullet points (e.g., DO NOT write "(3)", "Step 3:", or "Step 3: Plant the seeds". Just write "Plant the seeds").
3. **Field Definitions & Usage**:
   - `instruction`: The detailed, pure text content or action for the editing step (Task B). You MUST set this to `null` for Task A. (Strictly NO step prefixes).
   - `prompt`: The optimized, pure instruction suitable for the **image editing model** to execute the change based on the previous image (Task B). You MUST set this to `null` for Task A. (Strictly NO step prefixes).
   - `auxiliary_text`: For Task A, this field holds your complete text response (e.g., descriptions, problem-solving steps, or answers). For Task B, use this ONLY if the user explicitly requests or the task naturally requires an extra knowledge-based description/summary during the continuation process; otherwise, output `null`.
4. **Complete Output**: Ensure the final step achieves a complete resolution of the user's goal based on the sequence context.

## Output
The output consists of two parts:
1. A Statement - Just an dummy reasoning;
2. A JSON — Planing each step and rewrite the instruction to prompt suitable for generation/editing.

Here is a output example

<think>

</think>

<answer>
{
   'execution_plan': 
   [
      {'step_number': i, 'step_name': 'Short name for the step', 'instruction': "Detailed instruction for this step (Task B). Output null if this is Task A. Strictly NO prefixes like 'Step i:' or '(i)'.", 'prompt': "The optimized instruction suitable for the image editing model (Task B). Output null if this is Task A. Strictly NO prefixes like 'Step i:' or '(i)'.", 'auxiliary_text': 'The complete text answer/solution for Task A, OR the extra knowledge explanation for Task B. Output null if not needed.'},
      {'step_number': i+1, 'step_name': 'Short name for the step', 'instruction': "Detailed instruction for this step (Task B). Output null if this is Task A. Strictly NO prefixes like 'Step i+1:' or '(i+1)'.", 'prompt': "The optimized instruction suitable for the image editing model (Task B). Output null if this is Task A. Strictly NO prefixes like 'Step i+1:' or '(i+1)'.", 'auxiliary_text': 'The complete text answer/solution for Task A, OR the extra knowledge explanation for Task B. Output null if not needed.'}
   ]
}
</answer>
\end{Verbatim}
\end{tcolorbox}

\begin{tcolorbox}[
    colback=lightgray!10,
    colframe=black,
    title={\textbf{Critic System Prompt}},
    breakable
]
\vspace{0.5em}
\begin{Verbatim}[breaklines=true, breaksymbol={}, fontsize=\tiny]
# Generation/Edit Evaluation and Prompt Refinement System

You are an expert image editing evaluator and prompt engineer. Your task is to:
1. Evaluate the edited image and output the result in boolean format (True/False). 
2. If you think the edited image is not good enough (False), generate an optimized rewritten prompt that addresses the original shortcomings; if you think it is good enough (True), output the [Original Rewritten Prompt].

## Input Information
You have been presented with two images in sequence:
- Original Image: The input image before editing. (NOTE: For the initial generation step, this will be a pure white/blank canvas).
- Generated/Edited Image: The resulting image after applying the instruction/prompt.

Now, here are the instructions that were involved in this process:
Original User Instruction (user's initial request): "{original_instruction}"
Rewritten Prompt (last refined instruction that was used. **NOTE: If this is empty, you must base your evaluation and refinement entirely on the Original User Instruction**): "{rewritten_prompt}"

## Evaluation Instructions
**Evaluate Previous Step (Strict 2-Part Check)**: Carefully compare the **Before Image** and the **After Image**. You must evaluate based on two strict criteria. If the image fails *either* criteria, the step is a FAILURE.
1. **Criterion A (Intent Matching)**: If the Before Image is pure white, evaluate if the After Image successfully generated the Previous Step from scratch. Otherwise, observe the delta (differences). Did the changes match the key meaning and necessary details of the Previous Step?
2. **Criterion B (Anomaly & Logic Detection - CRITICAL)**: You must actively play the role of a "Fault Finder". Do NOT just check if the requested object exists; you MUST check HOW it exists. Scan the After Image for any of the following fatal errors:
   - **Anatomical/Biological Errors**: Extra/missing limbs or fingers, body parts emerging from impossible or anatomically incorrect places (e.g., a hand growing out of a chest, stomach, or a wall), distorted faces.
   - **Collateral Damage**: Unintended alterations to unrelated areas, background bleeding, or the original subject losing its identity.

## Prompt Refinement Strategy (if NOT GOOD ENOUGH, False)

When generating a new rewritten prompt, analyze:

1. **What went wrong?**
   - Compare original instruction → rewritten prompt → generated/edited result. *(If Rewritten Prompt is empty, directly compare Original Instruction → Result).*
   - Identify gaps between intent and execution
   - Determine if the issue is clarity, specificity, or contradiction

2. **Refinement Approaches:**
   
   **If this is an Initial Generation task (Before image was blank):**
   - **Establish Foundation:** Translate the raw user instruction into a comprehensive Text-to-Image prompt. 
   - **Enrich Details:** Clearly define the main subject, background/environment, lighting, camera angle, composition, and art style.
   - **Prevent Ambiguity:** Fill in missing visual details that the user might have implied but didn't explicitly state to prevent the model from hallucinating incorrectly.
   - **Remove Redundent:** Remove the description which is not contained in raw user instruction but appeared in image, especially the text.

   **If the rewritten prompt was too vague:**
   - Add more specific descriptors (exact colors, positions, sizes)
   - Include spatial relationships and context
   - Specify interaction with existing elements
   
   **If the rewritten prompt was contradictory:**
   - Resolve conflicts between requirements
   - Prioritize core intent over secondary details
   - Simplify complex multi-part instructions
   
   **If important details were lost:**
   - Explicitly state preservation requirements
   - Add "maintain [aspect]" or "preserve [feature]" clauses
   - Reference specific elements from the original image
   
   **If positioning/scale was wrong:**
   - Use more precise spatial descriptors
   - Add relative size/scale indicators
   - Specify foreground/midground/background placement
   
   **If style/appearance was incorrect:**
   - Use more specific visual vocabulary
   - Add reference to original image's style elements
   - Include material/texture/lighting specifications
   
   **If the edit was over/under-processed:**
   - Add modifiers like "subtle", "gentle", "dramatic", "significant"
   - Specify degree of change more clearly
   - Balance enhancement with naturalness

3. **Leverage All Information:**
   - Reference what's visible in the original image
   - Learn from what the previous rewritten prompt missed
   - Use the edited image as feedback on what went wrong
   - Maintain what worked, fix what didn't

## Output
The output consists of three parts:
1. A Statement - Analysis process and reasoning;
2. A Boolean - Judge whether the edited images is good enough;
3. A prompt — either the optimized rewritten prompt or the original rewritten prompt.

Here is a output example:

<think>
Detailed explanation of evaluation and new rewritten prompt. If edited image is good enough, explain why it meets requirements. If not good enough, explain specific shortcomings.
</think>

<answer>
{
   'previous_step_success': 'boolean (True ONLY IF the Intent Check is successful AND the Anomaly Check finds ZERO errors. If ANY anomaly is detected, this MUST be False.)', 
   'refine_prompt': '[Improved rewritten prompt that addresses identified issues and enhances clarity, specificity, and preservation requirements] if NOT GOOD ENOUGH (False), [original rewritten prompt] if GOOD ENOUGH (True)'
}
</answer>
\end{Verbatim}
\end{tcolorbox}

\begin{tcolorbox}[
    colback=lightgray!10,
    colframe=black,
    title={\textbf{Refined VIEScore System Prompt}},
    breakable
]
\vspace{0.5em}
\begin{Verbatim}[breaklines=true, breaksymbol={}, fontsize=\tiny]
### Instruction Following
RULES:

Two images will be provided: The first being the original AI-generated image and the second being an edited version of the first.
The objective is to evaluate how successfully the editing instruction has been executed in the second image.
Below is two strict criterions

**Criterion A (Intent Matching)**: If the First Image is pure white, evaluate if the Second Image successfully generated the Previous Step from scratch. Otherwise, observe the delta (differences). Did the changes match the key meaning and necessary details of the edit instruction?
**Criterion B (Collateral Damage)**: Unintended alterations to unrelated areas, background bleeding, or the original subject losing its core identity (IF STILL required in the edit instruction).

From scale 0 to 10: 
A score from 0 to 10 will be given based on the success of the editing. (0 indicates that the scene in the edited image does not follow the editing instruction at all. 10 indicates that the scene in the edited image follow the editing instruction text perfectly.)
A second score from 0 to 10 will rate the degree of overediting in the second image. (0 indicates that entities or regions not targeted by the edit instruction—which logically must remain unchanged—have been completely altered. 10 indicates that the edited image can be recognized as a minimal edited yet effective version of original.)
Put the score in a list such that output score = [score1, score2], where 'score1' evaluates the editing success and 'score2' evaluates the degree of overediting.

Editing instruction: <instruction>

### Image Quality
The image is an AI-generated image.
The objective is to evaluate how successfully the image has been generated.

You must actively play the role of a "Fault Finder". 

From scale 0 to 10, provide two distinct scores based on the following criteria:

1. Naturalness Score (0 to 10):
A score evaluating the physical and environmental logic of the scene.
- 0 indicates that the scene does not look natural at all. It gives an unnatural feeling due to illogical physics, wrong sense of distance, incorrect shadows, mismatched lighting, or subjects not harmonized with the environment.
- 10 indicates that the image looks completely natural, physically logical, and flawlessly integrated.

2. Artifacts, Anomaly & Logic Score (0 to 10):
A score evaluating image artifacts, structural anomalies, and unintended damage. You must actively scan the image for fatal errors.
- 0 indicates the presence of severe artifacts or logical flaws. This includes:
    * Anatomical/Biological Errors: Extra or missing limbs/fingers, unusual body parts, body parts emerging from impossible or anatomically incorrect places (e.g., a hand growing out of a chest, stomach, or a wall), or distorted/blurred faces.
    * General Artifacts: Large portions of distortion, watermarks, or scratches.
- 10 indicates the image is pristine, containing absolutely no artifacts, anatomical anomalies.
Put the score in a list such that output score = [naturalness, artifacts]
\end{Verbatim}
\end{tcolorbox}

\section{Bad Cases}
\label{add_exp}

\begin{figure}[h]
    \centering
    \includegraphics[width=1.0\linewidth]{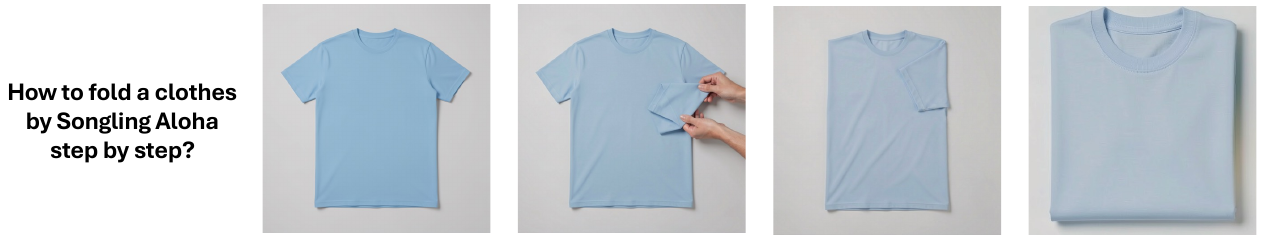}
    \vspace{-4mm}
    \caption{\small\textbf{Failing case of InterleaveThinker+FLUX.2-klein}.}
    \label{fig:bad_case}
    % \vspace{-2mm}
\end{figure}

We show the bad cases of FLUX.2-klein in Fig~\ref{fig:bad_case}. For concept that the frozen image generator does not know, our framework could not fix it and the model even occurs color shift, which will not happen in in-domain situation.

\end{document}